\documentclass{article}
\usepackage{spconf,amsmath,graphicx,multirow,subfigure}

\title{TreeMIL: a multi-instance learning framework for time series anomaly detection with inexact supervision}
\name{Chen Liu$^{\star}$ \qquad Shibo He$^{\star}$ \qquad Haoyu Liu$^{\dagger}$ \qquad Shizhong Li$^{\star}$}
\address{$^{\star}$ Zhejiang University \\
${\dagger}$NetEase Fuxi AI Lab}
\begin{document}
%
\maketitle
\begin{abstract}
Time series anomaly detection (TSAD) plays a vital role in various domains such as healthcare, networks and industry. Considering labels are crucial for detection but
difficult to obtain, we turn to TSAD with inexact supervision: only series-level labels are provided during the training phase, while point-level anomalies are predicted during the testing phase. 
Previous works follow a traditional multi-instance learning (MIL) approach, which focuses on encouraging high anomaly scores at individual time steps. However, time series anomalies are not only limited to individual point anomalies, they can also be collective anomalies, typically exhibiting abnormal patterns over subsequences. 
To address the challenge of collective anomalies, in this paper, we propose a tree-based MIL framework (TreeMIL).
We first adopt an $N$-ary tree structure to divide the entire series into multiple nodes, where nodes at different levels represent subsequences with different lengths. Then, the subsequences' features are extracted to determine the presence of collective anomalies. 
Finally, we calculate point-level anomaly scores by aggregating features from nodes at different levels.
Experiments conducted on seven public datasets and eight baselines demonstrate that TreeMIL achieves an average 32.3\% improvement in F1-score compared to previous state-of-the-art methods.
 The code is available at https://github.com/fly-orange/TreeMIL.


\end{abstract}
\section{Introduction}
\label{sec:intro}

\renewcommand{\thefootnote}{}
\footnotetext{This work was supported by the National Natural Science Foundation Program of China under Grant No. U23A20326 and No. U21B2029.}

Time series anomaly detection (TSAD) aims to identify abnormal time points or subsequences within the entire time series signals and has played critical roles in real-world applications like fault diagnosis, network intrusion detection, and health monitoring \cite{qin2023memory}. Previous studies mainly adopt a kind of unsupervised strategy, i.e., they use normal data exclusively to train generative models \cite{ruff2018deep,zhou2019beatgan,deng2021graph,xu2021anomaly,dai2022graph} and measure abnormality based on the generation error. However, such methodologies could potentially lead to more false alarms due to the limited variation in collected normal samples \cite{zhang2023exploiting}.
On the other hand, some other research, such as~\cite{wen2019time}, suggests incorporating anomaly labels during model training. While it can indeed improve the final results, from a practical perspective, collecting point-level labels is laborious and could be imprecise~\cite{zhang2023stad}, which will finally degrade the overall performance.

To address this problem, recent research has introduced a new paradigm, i.e., anomaly detection with inexact supervision~\cite{jiang2023weakly}. They suggest using coarse-grained anomaly labels during the training phase, such as series-level anomaly labels in time series rather than fine-grained point-level labels. This not only enhances detection performance and practicality but also aligns with the principles of Multi-Instance Learning (MIL).
In MIL, a collection of data instances constitutes a 'bag', where only the bag-level label is available, and the instance-level labels must be predicted \cite{sultani2018real}.  
While numerous MIL-based methods have emerged under this paradigm for video data \cite{chen2023mgfn,lv2023unbiased,li2022self}, there is a scarcity of solutions specifically designed for time series data. 
Janakiraman \emph{et~al.} \cite{janakiraman2018explaining} and Lee \emph{et~al.} \cite{lee2021weakly} apply conventional MIL to TSAD, which only encourages high anomaly scores at individual time steps, overlooking collective anomalies in the time series data. This flaw leads to high false negatives.


We hereby propose a tree-based MIL framework, TreeMIL. Our key objective is to enable the MIL structure to recognize both point and collective anomalies. To achieve this, we begin by formalizing the entire time series into an N-ary tree structure, where nodes represent subsequences of varying lengths. 
Next, the anomalous feature of each node is generated using an attention mechanism that incorporates information from its parent node, children nodes, neighbor nodes, and itself. Lastly, our anomaly discriminator considers anomaly features from subsequences at various scales and produces point-level anomaly scores, where information from multiple levels can be comprehensively incorporated.
Compared to unsupervised methods, our method is robust to noise because it captures the shared anomalous patterns.
We conduct experiments on seven public datasets and our method outperforms eight baselines. The contributions are summarized as follows:

\begin{figure*}[t]
\centering
\includegraphics[width=13cm]{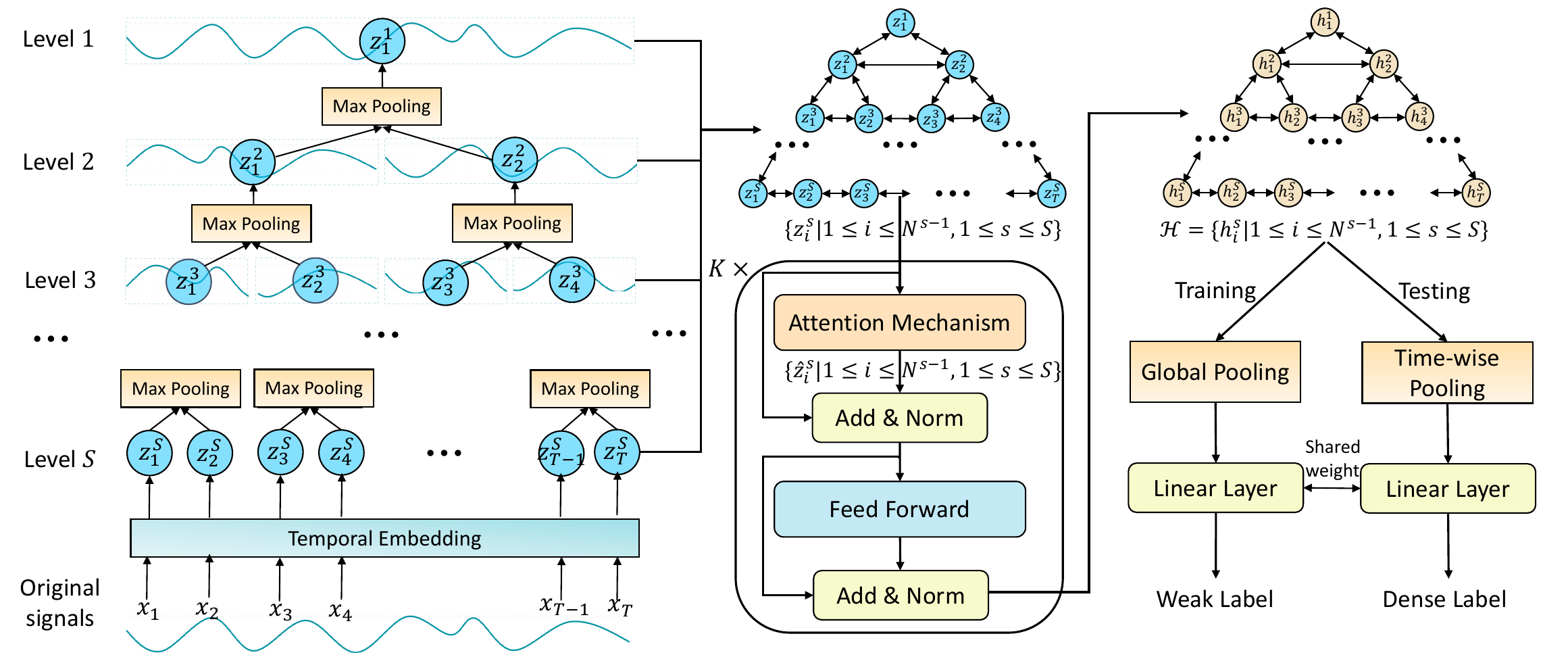}
\caption{Overall structure of the proposed TreeMIL framework. 
}
\label{fig:model}
\vspace{-0.6cm}
\end{figure*}

\begin{itemize}

\item We propose an MIL framework (TreeMIL) specially designed for time series data to achieve anomaly detection with inexact supervision. 

\item To model the collective anomalies that have been ignored by previous studies, we represent multi-scale subsequences within the entire time series as an $N$-ary tree structure, and find that incorporating this structure into MIL benefits identifying collective anomalies.

\item We conduct thorough experiments on seven public datasets to demonstrate the superiority of TreeMIL. The effectiveness of TreeMIL in identifying collective anomalies is experimentally verified through visualization results.

\end{itemize}



\section{Methodology}

\subsection{Problem Definition}
\label{ssec:prob}
Consider a $D$-dimension time series $X=[x_1,x_2,...,x_L] \in \mathbf{R}^{D \times L}$ of length $L$, where $x_t \in \mathbf{R}^{D}$ denotes the data collected at the $t$-th time step, 
we utilize a sliding window of length $T$ to split the entire series into non-overlapping time series segments, resulting in a dataset $\mathcal{D}=\left\{\mathbf{x}_1, \mathbf{x}_2, ...,\mathbf{x}_{M}  \right\}$, where $\mathbf{x}_i \in \mathbf{R}^{D \times T}$ denotes the $i$-th segment and $L=M \times T$. The dataset $\mathcal{D}$ is divided into non-overlapping training dataset $\mathcal{D}_{tr}$ and testing dataset $\mathcal{D}_{te}$. 

For the training dataset $\mathcal{D}_{tr}$, we annotate each subsequence with the occurrence of anomalies within it, resulting in $\left\{(\mathbf{x}_{i}, \mathbf{y}_{i}) | \mathbf{x}_{i} \in \mathcal{D}_{tr} \right\}$, where $\mathbf{y}_{i} \in \left\{0,1\right\} $ is the series-level label.
After training, given a sample $\mathbf{x}_{i} \in \mathcal{D}_{te}$, we produce point-level predictions $[\hat{y}_{i1},\hat{y}_{i2},...,\hat{y}_{iT}]$, where $\hat{y}_{it} \in \left\{0,1\right\} $ represents whether anomaly exists at the $t$-th time step in $\mathbf{x}_{i}$.

\subsection{Multi-resolution Temporal Embedding}
\label{ssec:const}
In this part, we create an ${N}$-ary tree using the input time series $\mathbf{x}_{i}$. In this tree, each parent node has $N$ children nodes. Fig.~\ref{fig:model} illustrates a special case of a binary tree where $N=2$. Specifically, each leaf node represents an individual time point, the root node symbolizes the entire time series, and the internal nodes signify subsequences of varying lengths. Consequently, nodes at higher levels encompass more time points. The tree has a total of $S=\lceil \log_NT \rceil+1$ levels, where $\lceil\rceil$ denotes the ceiling operation. 
The embedding of the $i$-th node at the $s$-th level is denoted as $z_{i}^{s}$ and calculated as follows.

First, we calculate the temporal embedding of each leaf node utilizing the uniform input representation method proposed in~\cite{zhou2021informer}, which includes a 1-D convolutional filter and a fixed position embedding. This transforms the input into $\mathbf{z}=[z_1^S,z_2^S,...,z_T^S] $, where $z_t^S$ represents the embedding of time point $t$. 
Then we compute the embeddings for all nodes in a bottom-up, level-by-level fashion, as depicted in Fig.~\ref{fig:model}. For instance, the embedding of the $i$-th node at the $s$-th level $z_{i}^{s}$ can be calculated by:
\begin{equation}
z_{i}^{s} = \mathbf{MaxPooling}(z_{(i-1)\times N+1}^{s+1},..., z_{i\times N}^{s+1}).
\end{equation}
where $z_{(i-1)\times N+1}^{s+1},..., z_{i\times N}^{s+1}$ are the embeddings of nodes at the ${s+1}$-th level.
To ensure the construction of a full $N$-ary tree, we pad the original time series to a length of $N^{S-1}$ with zeros before proceeding with the aforementioned procedures.

\subsection{Feature Extraction}
\label{ssec:fea}


Intuitively, a subsequence should be deemed anomalous when it contains anomalous time points or exhibits distinct patterns compared to its neighbor subsequences and the entire time series. Therefore, we calculate the features of each subsequence based on information from its children nodes, neighbor nodes, parent node, and itself, as shown in the middle part of Fig.~\ref{fig:model}. Specifically, for a node embedding $z_{i}^{s}$, we summarize the aforementioned node embeddings as a set:
\begin{equation}
\mathcal{N}_{i}^{s}=\left\{z_{i}^{s}, z_{(i-1)\times N+1}^{s+1},...,z_{i\times N}^{s+1}, z_{(i-\frac{l}{2})}^{s},...,z_{i+\frac{l}{2}}^{s},
z_{\frac{i}{N}}^{s-1}\right\}.
\end{equation}
where $l$ is a hyperparameter that controls the number of neighbor nodes.
Next, we calculate the features of $z_{i}^{s}$ using an attention mechanism:
\begin{equation}
\hat{z}_i^{s}=\sum_{\ell \in \mathcal{N}_{i}^{s}} \frac{\exp \left(q_i k_{\ell}^T / \sqrt{d_K}\right) v_{\ell}}{\sum_{\ell \in \mathcal{N}_{i}^{s}} \exp \left(q_i k_{\ell}^T / \sqrt{d_K}\right)}.
\end{equation}
where $q_i, k_i, v_i$ denote the query, key and value of the $i$-th node in the aforementioned set, with $d_K$ representing the dimension of these vectors. 
Note that, nodes that solely contain the padded values are excluded during the feature extraction process.
We also employ a multi-head mechanism, a feed-forward network and skip connections utilized in Transformer \cite{vaswani2017attention}. The number of attention layers is $K$.
Finally, we obtain $\mathcal{H}=\left\{ h_{i}^{s}|1\leq i \leq N^{s}, 1 \leq s \leq S \right\}$, where $h_{i}^{s} \in \mathbf{R}^{d}$ represents the anomaly features of the $i$-th node at the $s$-th level.

\subsection{Anomaly Discriminator}
\label{ssec:score}
During the training phase, to ensure that the feature of anomalies occurring at different scales can be learned, we place all node embeddings into a global pooling layer and generate a series-level label using a subsequent linear layer, as follows:
\begin{equation}
\hat{\mathbf{y}}= \sigma(W \times \mathbf{Pooling}(\mathcal{H}) + \mathbf{b}).
\end{equation}
where $W \in \mathbf{R}^{d \times 1}$ and $\mathbf{b} \in \mathbf{R}^{1}$ are the parameters of the linear layer, and $\sigma$ denotes the sigmoid function. We use binary cross-entropy as the loss function for training the model:
\begin{equation}
\mathcal{L}=-\sum_{(\mathbf{x},\mathbf{y}) \in \mathcal{D}_{tr}}[\mathbf{y} \cdot \log \hat{\mathbf{y}}+\left(1-\mathbf{y}\right) \cdot \log \left(1-\hat{\mathbf{y}}\right)].
\end{equation}
where $\mathbf{y}$ is the ground truth series-level anomaly label.
During the testing phase, to detect both point anomalies and collective anomalies, we propose a time-wise pooling mechanism. Specifically, for the $t$-th time point, we first aggregate the features of all nodes that encompass it:
\begin{equation}
\mathcal{H}_{t}=\left\{h_{t}^{S}, h_{\lceil\frac{t}{N}\rceil}^{S-1}, h_{\lceil\frac{t}{N^2}\rceil}^{S-2}..., h_{1}^{1}\right\}.
\end{equation}
then the prediction at the $t$-th time point is calculated by:
\begin{equation}
\hat{y_{t}}= \sigma(W \times \mathbf{Pooling}(\mathcal{H}_{t}) + \mathbf{b}).
\end{equation}
we finally obtain the predicted point-level anomaly labels $[\hat{y}_{1},\hat{y}_{2},...,\hat{y}_{T}]$ for the entire time series.

\section{Experiments}
\label{sec:exp}

\subsection{Experimental Setup}
\label{ssec: setup}

\textbf{Datasets.} We evaluate our method on seven public time series datasets spanning various domains.
\textbf{EMG}:
It records myographic signals from the subjects' forearms. 
\textbf{GECCO}: It collects time series data about drinking water composition. \textbf{SWAN-SF}: It contains space weather data from Harvard Dataverse. \textbf{Credit Card}: It records transactions of European cardholders.
\textbf{SMD} and \textbf{PSM}:
They contain metrics from server machines in an internet company. 
\textbf{SMAP}:
It contains telemetry data and the Incident Surprise Anomaly (ISA) reports.
Further details are presented in Table~\ref{tab:stat}. For EMG, we set $T$=500, following the approach in~\cite{lee2021weakly}. For GECCO, SWAN-SF, and Credit Card where anomalies persist for less than 40 time steps, we set $T=120$. For SMD and  PSM with anomalies exceeding 100 time steps, we set $T=720$. For SMAP with anomalies around 700 time steps, we set $T=1000$.

\textbf{Baselines.} We use two kinds of baselines.
\textbf{Unsupervised methods}: DeepSVDD~\cite{ruff2018deep}, BeatGAN~\cite{zhou2019beatgan}, GDN~\cite{deng2021graph}, AnoTrans~\cite{xu2021anomaly}, GANF~\cite{dai2022graph}.
\textbf{Weakly supervised methods}: DeepMIL~\cite{sultani2018real}, DTMIL~\cite{janakiraman2018explaining}, WETAS~\cite{lee2021weakly}.

\begin{table}[t]
\vspace{-0.3cm}
\caption{Details of datasets}\label{tab:stat}
\centering
\scalebox{0.8}{
\begin{tabular}{c|cccc}
\hline
Dataset       & Timestamps    &  Dimensions & Anomaly Ratio & Length \\  \hline
EMG     & 423825      & 8         & 5.97\%       &  500   \\
GECCO   & 138520      & 10         & 1.246\%       &  120   \\
SWAN-SF     & 120000      & 39         & 23.8\%       &  120   \\
Credit Card     & 284807      & 29         & 0.173\%       & 120   \\
SMD     & 708420   & 38      & 4.16\%     &  720    \\
PSM     & 87841    & 25     & 27.75\%     &   720   \\
SMAP    & 427617   & 25     & 12.79\%  &    1000 \\ \hline
\end{tabular}
}
\vspace{-0.5cm}
\end{table}

\textbf{Metrics.} We employ three metrics to evaluate the performance of all methods. \textbf{F1-W}: It denotes the F1 score between series-level labels and predictions. \textbf{F1-D}: It denotes the F1-score between point-level labels and predictions. \textbf{IoU}: It represents the intersection over union between point-level predictions and labels. 

\begin{table*}[t]
\centering
\caption{Performance comparison with baseline methods (without point-adjustment strategy)}\label{tab:result}
\scalebox{0.85}{
\begin{tabular}{c|ccccllcllcll}
\hline
\multirow{2}{*}{Methods} & \multicolumn{3}{c}{EMG}                                                    & \multicolumn{3}{c}{GECCO}                                                   & \multicolumn{3}{c}{SWAN-SF}                                                  & \multicolumn{3}{c}{Credit Card}                               \\ \cline{2-13} 
                         & F1-W                 & F1-D                     & IoU                      & F1-W                 & \multicolumn{1}{c}{F1-D} & \multicolumn{1}{c}{IoU} & F1-W                 & \multicolumn{1}{c}{F1-D} & \multicolumn{1}{c}{IoU} & F1-W                 & \multicolumn{1}{c}{F1-D} & IoU \\ \hline
DeepSVDD\cite{ruff2018deep}                 & -                    & 0.104                     & 0.054                     & -                    &     0.000                 &  0.000                   & -                    &     0.071                     &   0.037                      & -                    &       0.003                 &  0.002   \\
BeatGAN\cite{zhou2019beatgan}              & -                    & 0.105                     & 0.056                     & -                    &    0.000                  &    0.000                 & -                    &      0.069                    &    0.036                    & -                    &    0.003                     &  0.002   \\
GDN\cite{deng2021graph}                      & -                    & \multicolumn{1}{l}{0.192}     & 0.106     & -                    &      \underline{0.761}                    &    \underline{0.610}                    & -                    &     \underline{0.635}                     &    \underline{0.465}                    & -                    &     0.009                     &   0.004  \\ 
AnoTrans\cite{xu2021anomaly}                      & -                    & \multicolumn{1}{l}{0.106}     & \multicolumn{1}{l}{0.054}     & -                    &     0.070                     &     0.042                    & -                    &  0.019                        &   0.010                      & -                    &  0.007                        &  0.004   \\
GANF\cite{dai2022graph}                      & -                    & \multicolumn{1}{l}{0.151}     & 0.082     & -                    &     0.024                     &        0.012                & -                    &    0.483                      &   0.319                      & -                    &       0.005                 & 0.003   \\
\hline
DeepMIL\cite{sultani2018real}                  &  \underline{0.763}                    & 0.558                     & 0.387                     &  0.667    &     0.477                 &    0.313                 &   0.934     &        0.555                &     0.384                    & \multicolumn{1}{l}{0.846} &  0.220                        &  0.124   \\
DTMIL \cite{janakiraman2018explaining}                    & 0.562 & 0.016     & 0.008      & 0.560 &   0.142                       &    0.076                    & \underline{0.940}  &       0.504           &     0.337                & \multicolumn{1}{l}{0.841} &  0.290                       &  0.170  \\
WETAS\cite{lee2021weakly}                    &  0.752                    & \underline{0.580}                     & \underline{0.412}                     & \underline{0.750} & 0.551                    & 0.310                    & 0.887 & 0.561                     & 0.390                    & \underline{0.859}  &   \textbf{0.376}                       &  \textbf{0.231}   \\
TreeMIL(ours)            & \textbf{0.818} & \textbf{0.672} & \textbf{0.502} & \textbf{1.000}    & \textbf{0.860}                     & \textbf{0.755}                  & \textbf{0.948}  & \textbf{0.670}                     & \textbf{0.504}                    & \textbf{0.867} & \underline{0.337}                         & \underline{0.203}     \\ \hline
\multirow{2}{*}{Methods} & \multicolumn{3}{c}{SMD}                                                    & \multicolumn{3}{c}{PSM}                                                   & \multicolumn{3}{c}{SMAP}                                                  & \multicolumn{3}{c}{Avg}                               \\ \cline{2-13} 
                         & F1-W                 & F1-D                     & IoU                      & F1-W                 & \multicolumn{1}{c}{F1-D} & \multicolumn{1}{c}{IoU} & F1-W                 & \multicolumn{1}{c}{F1-D} & \multicolumn{1}{c}{IoU} & F1-W                 & \multicolumn{1}{c}{F1-D} & IoU \\ \hline
DeepSVDD\cite{ruff2018deep}                 & -                    & 0.014                    & 0.008                     & -                    & 0.218                     & 0.122                    & -                    &       0.028                   &   0.015                      & -                    &  0.063                        & 0.034    \\
BeatGAN\cite{zhou2019beatgan}              & -                    & 0.122                     & 0.066                     & -                    & 0.253                     & 0.144                    & -                    & 0.033                         &   0.017                      & -                    &     0.084                    &  0.046   \\
GDN\cite{deng2021graph}                      & -                    & \multicolumn{1}{l}{0.134}     & 0.072     & -                    &     \underline{0.595}                     &   \underline{0.423}                     & -                    &    0.302                      &   0.144                      & -                    &        0.375                  & 0.261    \\ 
AnoTrans\cite{xu2021anomaly}                      & -                    & \multicolumn{1}{l}{0.037}     & \multicolumn{1}{l}{0.019}     & -                    & 0.013                         &  0.007                       & -                    &     0.019                     &  0.010                       & -                    &     0.039                     &  0.021   \\
GANF\cite{dai2022graph}                      & -                    & \multicolumn{1}{l}{0.065}     & 0.033     & -                    &      0.561                    &  0.390                      & -                    &     0.068                     &  0.035                       & -                    &   0.194                       & 0.125    \\
\hline
DeepMIL\cite{sultani2018real}                  &  0.400                    & 0.163                     & 0.089                     & \textbf{0.800} & 0.414                     & 0.261                    & 0.514 &  0.361                        &  0.220                       & 0.703  & 0.393                       &   0.254  \\
DTMIL \cite{janakiraman2018explaining}                    & \textbf{0.800} & 0.156     & 0.085      & 0.743 &  0.163                        &   0.089                      &  0.454 &   0.022                    &    0.011                     &  0.700   &   0.185                       &  0.111   \\
WETAS\cite{lee2021weakly}                    &  0.602                    & \underline{0.294}                     & \underline{0.207}                     & 0.741 & 0.481                     & 0.317                    & \underline{0.568} &  \underline{0.386}                     &  \underline{0.239}                    & \underline{0.737}   &   \underline{0.461}                       &  \underline{0.301}   \\
TreeMIL(ours)            & \underline{0.769} & \textbf{0.522} & \textbf{0.353} & \underline{0.764} & \textbf{0.715}                     & \textbf{0.557}                    & \textbf{0.675} & \textbf{0.494}                     & \textbf{0.328}                    & \textbf{0.834} & \textbf{0.610}                         & \textbf{0.457}     \\ \hline
\end{tabular}}
\vspace{-0.6cm}
\end{table*}

\textbf{Implementation.}
The baselines are implemented based on the hyperparameters reported in previous literature. For our model, we set $N=2$, indicating the construction of a binary tree. As for the attention mechanism, we use $l=3$, $K=2$, and $d_K=d=128$. During the training stage, we utilize an Adam optimizer with a learning rate of 0.0001 and a batch size of 32. All our experiments are conducted on a single RTX 3090.

\subsection{Experiment Result and Analysis}
\label{ssec: result}

\textbf{Performance comparison.}
Results are reported in Table ~\ref{tab:result}
Because using the point-adjustment strategy will result in artificially inflated metrics \cite{doshi2022reward}, \textbf{the results of all methods are calculated without this strategy}. The best and the second-best results are highlighted in bold and underlined,  respectively. Our method achieves an average improvement of 32.3\% in F1-D and 51.8\% in IoU compared to previous SOTA methods.
WETAS performs slightly better than TreeMIL on the Credit Card dataset because the dataset contains only point anomalies. For other datasets containing both point and collective anomalies, our method consistently achieves the best performance. 
Additionally, three observations can be made: Firstly, most unsupervised methods exhibit inferior performance compared to weakly-supervised methods, indicating that coarse-grained labels benefit anomaly detection. Secondly, the prediction-based method (GDN) outperforms other unsupervised methods, implying that the prediction task is more suitable for anomaly detection. Thirdly, among weakly supervised methods, DTMIL performs the worst because it tries to maximize the anomaly score of individual points. While DeepMIL and WETAS consider the outlyingness of subsequences, they calculate it by summarizing the anomaly score of time points, which ignores patterns of subsequences. TreeMIL overcomes this drawback by representing multi-scale subsequences in an $N$-ary tree structure. 

\begin{figure}[t]
\centering
\subfigure[EMG]{
\includegraphics[width=4.07cm]{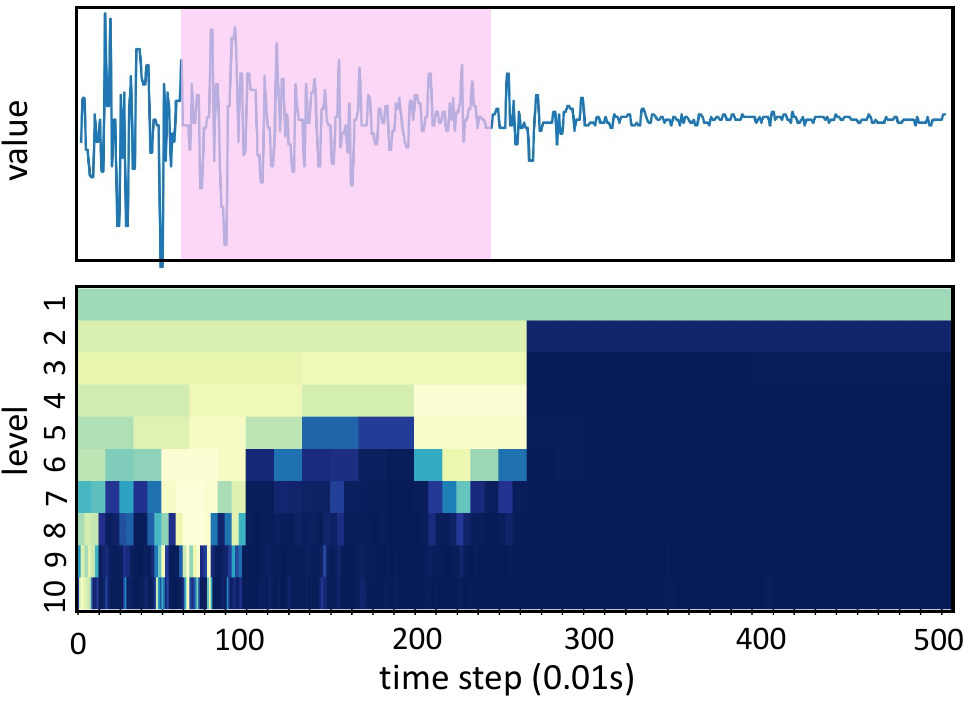}
\label{fig:case1}
}%
\subfigure[GECCO]{
\includegraphics[width=4.66cm]{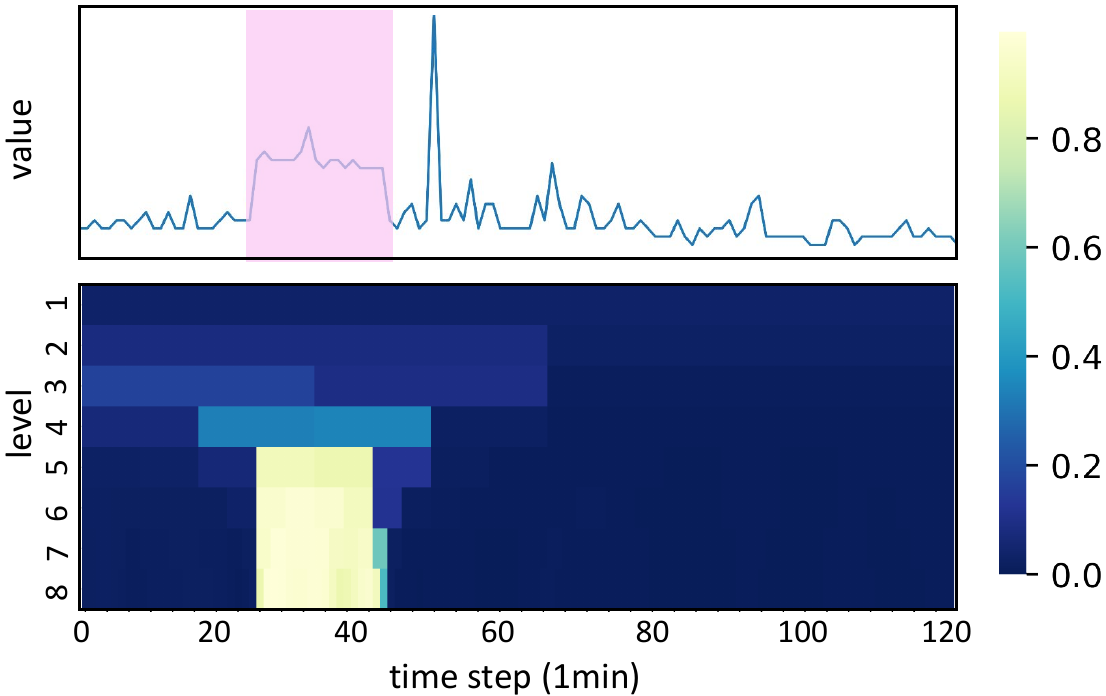}
\label{fig:case2}
}%
\caption{\textbf{Anomaly score map}: The x-axis represents the time step and the y-axis represents the level of the tree. The red rectangle denotes the labeled anomaly window.}
\label{fig:map}
\vspace{-0.6cm}
\end{figure}

\textbf{Anomaly score visualization.}
To illustrate how subsequences at different scales contribute to the final point-level predictions, we present the anomaly scores of all nodes in a heat map. As shown in Fig.~\ref{fig:case1}, for a long collective anomaly, the nodes at lower levels give higher scores, and those at higher levels could avoid giving high scores. In the case of a short collective anomaly (Fig.~\ref{fig:case2}), the high anomaly scores are assigned by nodes at higher levels. This demonstrates that collective anomalies of varying lengths can be effectively captured by multi-scale subsequences.

\begin{figure}[t]
\centering
\subfigure[Pooling type]{
\includegraphics[width=4.28cm]{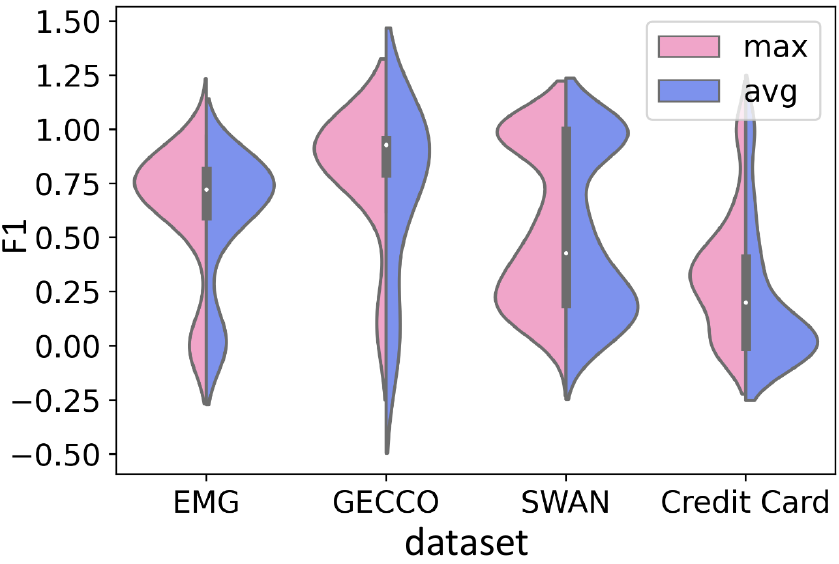}
\label{fig:pool}
}%
\subfigure[Ary size]{
\includegraphics[width=4.3cm]{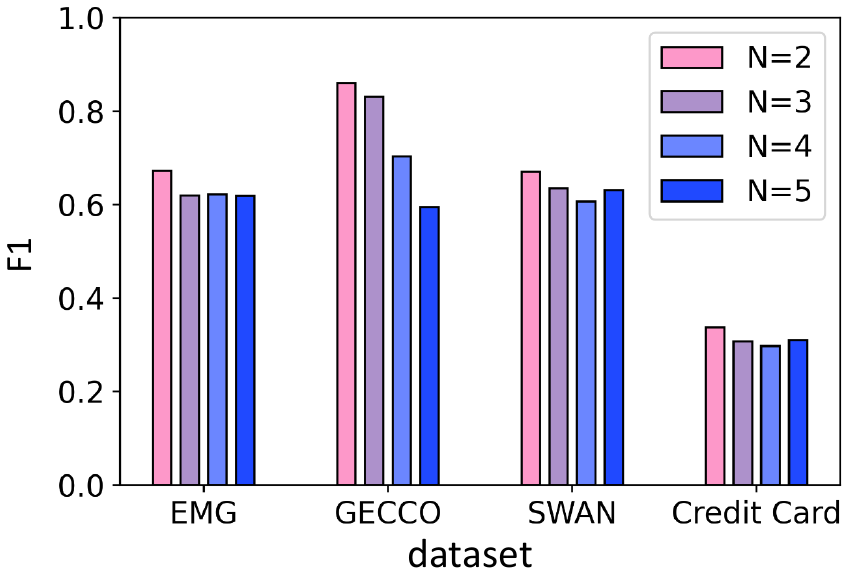}
\label{fig:ary}
}%
\caption{\textbf{Sensitivity analysis}: a) The violet plot describes the F1-D distribution of all time series when applying max and average pooling. b) The bar plot describes the F1-score when applying different ary size.}
\vspace{-0.6cm}
\end{figure}

\textbf{Sensitivity analysis}.
We also conduct a sensitivity analysis of pooling type and ary size $N$. 
Fig.~\ref{fig:pool} shows that the two types have a minimal impact on EMG, whereas max pooling outperforms average pooling on GECCO, SWAN, and PSM because these datasets contain more short collective anomalies. It is easier for max pooling to capture short anomalous subsequences. Fig.~\ref{fig:ary} shows that setting $N=2$ provides the best results since it captures more fine-grained temporal information. Different sizes help detect anomalies at different scales, leading to similar results, except for GECCO, where the anomaly length is relatively fixed. 



\section{Conclusion}
\label{sec:conclu}
In this paper, we propose TreeMIL for TSAD with inexact supervision. To address the challenge of collective anomalies, which have been overlooked by previous studies, we represent the entire time series using an $N$-ary tree structure. This structure is seamlessly integrated into the MIL framework, where we aggregate the feature from multi-scale nodes to produce final point-level predictions. Only series-level labels are provided during the training phases. Experiments conducted on seven public datasets demonstrate that TreeMIL outperforms all unsupervised and weakly supervised methods.

\vfill\pagebreak

\bibliographystyle{IEEEbib}
\bibliography{refs}

\begin{thebibliography}{10}
\providecommand{\url}[1]{#1}
\csname url@samestyle\endcsname
\providecommand{\newblock}{\relax}
\providecommand{\bibinfo}[2]{#2}
\providecommand{\BIBentrySTDinterwordspacing}{\spaceskip=0pt\relax}
\providecommand{\BIBentryALTinterwordstretchfactor}{4}
\providecommand{\BIBentryALTinterwordspacing}{\spaceskip=\fontdimen2\font plus
\BIBentryALTinterwordstretchfactor\fontdimen3\font minus \fontdimen4\font\relax}
\providecommand{\BIBforeignlanguage}[2]{{%
\expandafter\ifx\csname l@#1\endcsname\relax
\typeout{** WARNING: IEEEtran.bst: No hyphenation pattern has been}%
\typeout{** loaded for the language `#1'. Using the pattern for}%
\typeout{** the default language instead.}%
\else
\language=\csname l@#1\endcsname
\fi
#2}}
\providecommand{\BIBdecl}{\relax}
\BIBdecl

\bibitem{qin2023memory}
S.~Qin, Y.~Luo, and G.~Tao, ``Memory-augmented u-transformer for multivariate time series anomaly detection,'' in \emph{ICASSP 2023-2023 IEEE International Conference on Acoustics, Speech and Signal Processing (ICASSP)}.\hskip 1em plus 0.5em minus 0.4em\relax IEEE, 2023, pp. 1--5.

\bibitem{jiang2023weakly}
M.~Jiang, C.~Hou, A.~Zheng, X.~Hu, S.~Han, H.~Huang, X.~He, P.~S. Yu, and Y.~Zhao, ``Weakly supervised anomaly detection: A survey,'' \emph{arXiv preprint arXiv:2302.04549}, 2023.

\bibitem{wen2019time}
T.~Wen and R.~Keyes, ``Time series anomaly detection using convolutional neural networks and transfer learning,'' \emph{arXiv preprint arXiv:1905.13628}, 2019.

\bibitem{zhou2019beatgan}
B.~Zhou, S.~Liu, B.~Hooi, X.~Cheng, and J.~Ye, ``Beatgan: Anomalous rhythm detection using adversarially generated time series.'' in \emph{IJCAI}, vol. 2019, 2019, pp. 4433--4439.

\bibitem{xu2021anomaly}
J.~Xu, H.~Wu, J.~Wang, and M.~Long, ``Anomaly transformer: Time series anomaly detection with association discrepancy,'' \emph{arXiv preprint arXiv:2110.02642}, 2021.

\bibitem{lee2021weakly}
D.~Lee, S.~Yu, H.~Ju, and H.~Yu, ``Weakly supervised temporal anomaly segmentation with dynamic time warping,'' in \emph{Proceedings of the IEEE/CVF International Conference on Computer Vision}, 2021, pp. 7355--7364.

\bibitem{sultani2018real}
W.~Sultani, C.~Chen, and M.~Shah, ``Real-world anomaly detection in surveillance videos,'' in \emph{Proceedings of the IEEE conference on computer vision and pattern recognition}, 2018, pp. 6479--6488.

\bibitem{li2022self}
S.~Li, F.~Liu, and L.~Jiao, ``Self-training multi-sequence learning with transformer for weakly supervised video anomaly detection,'' in \emph{Proceedings of the AAAI Conference on Artificial Intelligence}, vol.~36, no.~2, 2022, pp. 1395--1403.

\bibitem{chen2023mgfn}
Y.~Chen, Z.~Liu, B.~Zhang, W.~Fok, X.~Qi, and Y.-C. Wu, ``Mgfn: Magnitude-contrastive glance-and-focus network for weakly-supervised video anomaly detection,'' in \emph{Proceedings of the AAAI Conference on Artificial Intelligence}, vol.~37, no.~1, 2023, pp. 387--395.

\bibitem{janakiraman2018explaining}
V.~M. Janakiraman, ``Explaining aviation safety incidents using deep temporal multiple instance learning,'' in \emph{Proceedings of the 24th ACM SIGKDD International Conference on Knowledge Discovery \& Data Mining}, 2018, pp. 406--415.

\bibitem{zhou2021informer}
H.~Zhou, S.~Zhang, J.~Peng, S.~Zhang, J.~Li, H.~Xiong, and W.~Zhang, ``Informer: Beyond efficient transformer for long sequence time-series forecasting,'' in \emph{Proceedings of the AAAI conference on artificial intelligence}, vol.~35, no.~12, 2021, pp. 11\,106--11\,115.

\bibitem{deng2021graph}
A.~Deng and B.~Hooi, ``Graph neural network-based anomaly detection in multivariate time series,'' in \emph{Proceedings of the AAAI conference on artificial intelligence}, vol.~35, no.~5, 2021, pp. 4027--4035.

\bibitem{doshi2022reward}
K.~Doshi, S.~Abudalou, and Y.~Yilmaz, ``Reward once, penalize once: Rectifying time series anomaly detection,'' in \emph{2022 International Joint Conference on Neural Networks (IJCNN)}.\hskip 1em plus 0.5em minus 0.4em\relax IEEE, 2022, pp. 1--8.

\bibitem{dai2022graph}
E.~Dai and J.~Chen, ``Graph-augmented normalizing flows for anomaly detection of multiple time series,'' \emph{arXiv preprint arXiv:2202.07857}, 2022.

\bibitem{ruff2018deep}
L.~Ruff, R.~Vandermeulen, N.~Goernitz, L.~Deecke, S.~A. Siddiqui, A.~Binder, E.~M{\"u}ller, and M.~Kloft, ``Deep one-class classification,'' in \emph{International conference on machine learning}.\hskip 1em plus 0.5em minus 0.4em\relax PMLR, 2018, pp. 4393--4402.

\bibitem{vaswani2017attention}
A.~Vaswani, N.~Shazeer, N.~Parmar, J.~Uszkoreit, L.~Jones, A.~N. Gomez, {\L}.~Kaiser, and I.~Polosukhin, ``Attention is all you need,'' \emph{Advances in neural information processing systems}, vol.~30, 2017.

\bibitem{zhang2023exploiting}
C.~Zhang, G.~Li, Y.~Qi, S.~Wang, L.~Qing, Q.~Huang, and M.-H. Yang, ``Exploiting completeness and uncertainty of pseudo labels for weakly supervised video anomaly detection,'' in \emph{Proceedings of the IEEE/CVF Conference on Computer Vision and Pattern Recognition}, 2023, pp. 16\,271--16\,280.

\bibitem{lv2023unbiased}
H.~Lv, Z.~Yue, Q.~Sun, B.~Luo, Z.~Cui, and H.~Zhang, ``Unbiased multiple instance learning for weakly supervised video anomaly detection,'' in \emph{Proceedings of the IEEE/CVF Conference on Computer Vision and Pattern Recognition}, 2023, pp. 8022--8031.

\bibitem{zhang2023stad}
Z.~Zhang, W.~Li, W.~Ding, L.~Zhang, Q.~Lu, P.~Hu, T.~Gui, and S.~Lu, ``Stad-gan: unsupervised anomaly detection on multivariate time series with self-training generative adversarial networks,'' \emph{ACM Transactions on Knowledge Discovery from Data}, vol.~17, no.~5, pp. 1--18, 2023.

\end{thebibliography}

\end{document}